\documentclass[sigconf]{acmart}

\AtBeginDocument{%
  }

\copyrightyear{2025}
\acmYear{2025}
\setcopyright{cc}
\setcctype{by}\setcctype{by}
\acmConference[ICMI Companion '25]{Companion Proceedings of the 27th
International Conference on Multimodal Interaction}{October 13--17,
2025}{Canberra, ACT, Australia}
\acmBooktitle{Companion Proceedings of the 27th International Conference on
Multimodal Interaction (ICMI Companion '25), October 13--17, 2025, Canberra,
ACT, Australia}
\acmDOI{10.1145/3747327.3764788}
\acmISBN{979-8-4007-2076-5/2025/10}

\usepackage[inline]{enumitem}
\usepackage{algorithmic}
\usepackage{graphicx}
\usepackage{textcomp}
\usepackage{xcolor}

\usepackage{witharrows}

\usepackage[flushleft]{threeparttable}
\usepackage{tablefootnote}
\usepackage{array}
\newcolumntype{P}[1]{>{\centering\arraybackslash}p{#1}}
\usepackage{multicol}
\usepackage{multirow}
\usepackage{tabulary}
\usepackage{colortbl}
\definecolor{mygray}{gray}{0.90}
\usepackage{makecell}
\usepackage{booktabs}
\usepackage{subcaption}

\usepackage{stmaryrd}

\usepackage{float}
\usepackage{pifont}

\usepackage{arydshln}
\usepackage[switch,columnwise]{lineno}

\begin{document}

\title{Tiny-BioMoE: a Lightweight Embedding Model for Biosignal Analysis}

\author{Stefanos Gkikas}
\email{gkikas@ics.forth.gr}
\orcid{0000-0002-4123-1302}
\affiliation{%
  \institution{Foundation for Research \& Technology-Hellas}
  \city{Heraklion}
  \country{Greece}
}

\author{Ioannis Kyprakis}
\email{ikyprakis@ics.forth.gr}
\orcid{0009-0000-0711-5108}
\affiliation{%
  \institution{Foundation for Research \& Technology-Hellas}
  \city{Heraklion}
  \country{Greece}
}

\author{Manolis Tsiknakis}
\email{tsiknaki@ics.forth.gr}
\orcid{0000-0001-8454-1450}
\affiliation{%
  \institution{Foundation for Research \& Technology-Hellas and Hellenic Mediterranean University}
  \city{Heraklion}
  \country{Greece}
}


\begin{abstract}
Pain is a complex and pervasive condition that affects a significant portion of the population. Accurate and consistent assessment is essential for individuals suffering from pain, as well as for developing effective management strategies in a healthcare system.
Automatic pain assessment systems enable continuous monitoring, support clinical decision-making, and help minimize patient distress while mitigating the risk of functional deterioration. Leveraging physiological signals offers objective and precise insights into a person's state, and their integration in a multimodal framework can further enhance system performance.
This study has been submitted to the \textit{Second Multimodal Sensing Grand Challenge for Next-Gen Pain Assessment (AI4PAIN)}. 
The proposed approach introduces \textit{Tiny-BioMoE}, a lightweight pretrained embedding model for biosignal analysis. Trained on $4.4$ million biosignal image representations and consisting of only $7.3$ million parameters, it serves as an effective tool for extracting high-quality embeddings for downstream tasks. Extensive experiments involving electrodermal activity, blood volume pulse, respiratory signals, peripheral oxygen saturation, and their combinations highlight the model's effectiveness across diverse modalities in automatic pain recognition tasks.
\textit{\textcolor{blue}{The model's architecture (code) and weights are available at \url{https://github.com/GkikasStefanos/Tiny-BioMoE}.}}

\end{abstract}




\maketitle

\section{Introduction}
Pain serves as a vital evolutionary mechanism, alerting the organism to potential harm or signaling the onset of illness, and plays a vital role in the body's defense system by helping maintain physiological integrity \cite{santiago_2022}. It is a subjective experience comprising multiple components, including nociceptive, sensory, affective, and cognitive dimensions \cite{marchand_2024}.
Pain has been described as a \textit{\textquotedblleft Silent Public Health Epidemic\textquotedblright} \cite{katzman_gallagher_2024}, emphasizing its widespread yet often underestimated impact. In nursing literature, it is also referred to as \textit{\textquotedblleft the fifth vital sign\textquotedblright} \cite{joel_lucille_1999}, reflecting the need for its routine and systematic assessment alongside other vital signs.
In addition, opioid analgesics are the most frequently prescribed treatment for pain management \cite{kaye_jones_2017}, yet they often lead to addiction and overdose \cite{stampas_pedroza_2020}. Moreover, their side effects—such as lethargy, depression, anxiety, and nausea—significantly affect both workforce productivity and overall quality of life \cite{benyamin_trescot_2008}.
The inherent subjectivity and complexity of pain assessment have been identified as significant challenges in both research and clinical practice. For example, pain management often relies on patients' subjective reports, making it difficult to administer medication with precision. This lack of objective evaluation contributes significantly to the overprescription and overuse of pain medications \cite{kong_chon_2024}.
Moreover, managing and assessing pain in patients with--or at risk of--medical instability presents significant clinical challenges, particularly when communication barriers are present \cite{puntilo_staannard_2022}. Research highlights that pain in critically ill adults remains frequently under-managed. A significant limitation is the lack of structured, comprehensive tools for assessing pain and supporting clinical decision-making in these contexts \cite{meehan_mcrae_1995}. Furthermore, cancer-related pain is highly prevalent, particularly in advanced stages of the disease, with its incidence exceeding $40\%$ \cite{bang_hak_2023}. Additionally, the presence of negative emotions such as fear and sadness can influence pain expression, further complicating accurate assessment \cite{tessier_mazet_2024}.

Precise evaluation and understanding of the factors influencing pain are crucial for achieving effective pain management \cite{sabbadini_massaroni_2020}.
Self-report methods, such as numerical rating scales and questionnaires, remain the gold standard for evaluating patient experiences. However, their reliability significantly diminishes in cases where patients exhibit altered consciousness, cognitive impairments, or communication difficulties \cite{herr_coyne_2006}.
Behavioral cues—such as facial expressions, vocalizations, and body movements—are widely used to infer pain, particularly in patients who cannot communicate \cite{rojas_brown_2023}. Complementary to these, physiological signals like electrocardiography (ECG), electromyography (EMG), and electrodermal activity (EDA) provide deeper insights into the body’s response to pain \cite{gkikas_tsiknakis_slr_2023}.
Physiological signals play a crucial role in pain assessment by enabling a more objective and accurate understanding of a person’s condition. Sabbadini \textit{et al}. \cite{sabbadini_tomaso_2024} emphasized the importance of incorporating raw physiological data, such as continuous waveforms, to capture the intricate relationship between pain and physiological responses. Gozzi \textit{et al}. \cite{gozzi_preatoni_2024} advocated for a broader perspective on pain as a multidimensional phenomenon, emphasizing the integration of physiological biomarkers with psychosocial factors.

This study introduces a lightweight embedding model, pretrained on over $4$ million biosignals, and applies it to automatic pain assessment tasks across a wide range of modalities. Although numerous studies rely on biosignals, few models leverage large-scale pretraining. Moreover, given the growing concerns around the computational cost of large models for both training and inference, this work proposes a compact and efficient alternative, aiming to ensure accessibility regardless of the user’s hardware capabilities.

\section{Related Work}
Over the past decade, automatic pain assessment has gained increasing attention, with advancements shifting from conventional image and signal processing methods to more complex deep learning-based techniques \cite{gkikas_phd_thesis_2025}. The majority of existing methods are video-based, aiming to capture behavioral cues through facial expressions, body movements, or other visual indicators and employing a wide range of modeling strategies \cite{gkikas_tsiknakis_embc, bargshady_hirachan_2024,gkikas_tsiknakis_thermal_2024,huang_dong_2022}.
While video-based approaches dominate the field, a considerable number of studies have also focused on biosignal-based methods, although to a lesser extent. These works have investigated the utility of various physiological signals, such as electrocardiography (ECG) \cite{gkikas_chatzaki_2022, gkikas_chatzaki_2023}, electromyography (EMG) \cite{pavlidou_tsiknakis_2025,patil_patil_2024,thiam_bellmann_kestler_2019,werner_hamadi_niese_2014}, electrodermal activity (EDA) \cite{aziz_joseph_2025,li_luo_2024,lu_ozek_kamarthi_2023,phan_iyortsuun_2023,ji_zhao_li_2023}, and brain activity through functional near-infrared spectroscopy (fNIRS) \cite{rojas_huang_2016, rojas_liao_2019,rojas_romero_2021,rojas_joseph_bargshady_2024,khan_sousani_2024,bargshady_aziz_2025}. For a more comprehensive analysis of biosignal modalities within automatic pain recognition frameworks, the reader is referred to \cite{khan_umar_2025}.

In addition, multimodal approaches combining behavioral and physiological data have gained increasing attention in recent years, with several studies demonstrating the benefits of integrating multiple sources of information to improve performance \cite{farmani_bargshady_2025}.
Many researchers explore combination of biosignals due to their potential to provide more accurate measurements.
For instance, Jiang \textit{et al}. \cite{jiang_li_he_2024} proposed a hybrid temporal-channel attention model that fuses ECG and galvanic skin response (GSR) signals, while in \cite{jiang_rosio_2024}, the authors employed ECG and GSR signals alongside neural networks enhanced with Squeeze-and-Excitation blocks. Chu \textit{et al}. \cite{chu_zhao_2017} integrated blood volume pulse, ECG, and skin conductance level, using a hybrid approach that combined genetic algorithm-based feature selection with principal component analysis.
On the other hand, Badura \textit{et al}. \cite{badura_2021} employed a multimodal setup in their effort to assess pain during physiotherapy, which included EDA, EMG, respiration, blood volume pulse (BVP), and hand force.
Other researchers focus on combining physiological with behavioral modalities. The authors in \cite{zhi_yu_2019} utilized facial videos in combination with ECG, EMG, and EDA, extracting a wide range of handcrafted features and fusing them.  In \cite{gkikas_tachos_2024}, the authors combined facial videos and heart rate data to develop a transformer-based framework for recognizing pain intensity. Similarly, in \cite{farmani_giuseppi_2025}, CNN and LSTM models were used to analyze facial videos and EDA, achieving high performance.
Finally, the authors in \cite{gkikas_rojas_painformer_2025} proposed a foundation model that extracted feature representations from a wide range of modalities—both behavioral and physiological—for pain assessment.

\section{Methodology}
This section outlines the architecture of the proposed embedding model, the pretraining procedure, and the dataset used for pretraining. It also details the biosignal preprocessing steps and the transformation of these signals into visual representations. Additionally, it describes the augmentation and regularization techniques employed for the pain recognition tasks.

\subsection{Tiny-BioMoE}
\begin{figure}
\begin{center}
\includegraphics[scale=0.75]{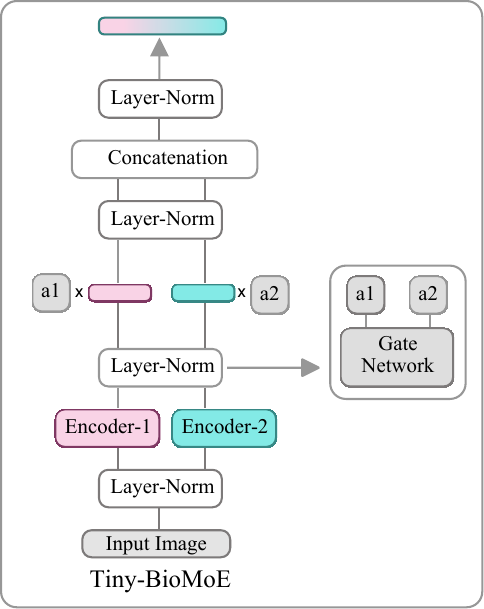} 
\end{center}
\caption{Overview of the Tiny-BioMoE architecture.}
\label{tiny}
\end{figure}

Inspired by recent advancements in deep learning involving Mixture of Experts (MoE) architectures for language \cite{cai_jiang_2025} and vision tasks \cite{riquelme_2021}, we propose \textit{Tiny-BioMoE}---a lightweight MoE model comprising only $7.34$ million parameters. The model consists of two vision transformer encoders, \textit{Encoder-1} and \textit{Encoder-2}.
Both encoders process the input image independently to extract their respective embedding representations, which are subsequently fused into a unified feature vector.
Table \ref{table:module_parameters} reports the number of parameters and computational cost, measured in floating-point operations (FLOPs), while Figure  \ref{tiny} provides a high-level architectural overview.

\subsubsection{Encoder-1}
Integrates spectral and self-attention mechanisms in the initial layers, while relying exclusively on self-attention in the later stages.
Each input image $I$ is divided into $n$ non-overlapping $16 \times 16$ patches, reshaped into tokens $\in \mathbb{R}^{n \times d}$ with $d = 768$, followed by positional encoding.
To incorporate frequency information, a 2D Fast Fourier Transform (FFT) is applied to each token $x$, yielding $X = \mathscr{F}[x] \in \mathbb{C}^{h \times w \times d}$. A learnable complex-valued filter $K \in \mathbb{C}^{h \times w \times d}$ modulates the spectral components via element-wise multiplication: $\tilde{X} = K \odot X$. The filtered signal is then transformed back into the spatial domain using the inverse FFT: $x \leftarrow \mathscr{F}^{-1}[\tilde{X}]$.
A depthwise convolution-based MLP enhances channel-wise interactions, 
$f(x) = W_2 \cdot \text{GELU}(\text{DWConv}(W_1 \cdot x + b_1)) + b_2$,
with layer normalization applied before and after the FFT and IFFT.
The attention layer employs standard scaled dot-product self-attention:
\begin{equation}
\widetilde{X} = \text{Attn}(X W_q, X W_k, X W_v),
\end{equation}
where $W_q$, $W_k$, and $W_v \in \mathbb{R}^{d \times d}$ are the projection matrices for queries, keys, and values, respectively.
\textit{Encoder-1} is composed of four hierarchical stages, each reducing spatial resolution by a factor of 2. The embedding dimensions across the stages are $64$, $128$, $320$, and $96$, with $1$ attention head used per stage.
\hyperref[encoders]{Figure 2(a}) illustrates the architecture of \textit{Encoder-1}.

\subsubsection{Encoder-2}
\begin{figure}
\begin{center}
\includegraphics[scale=0.57]{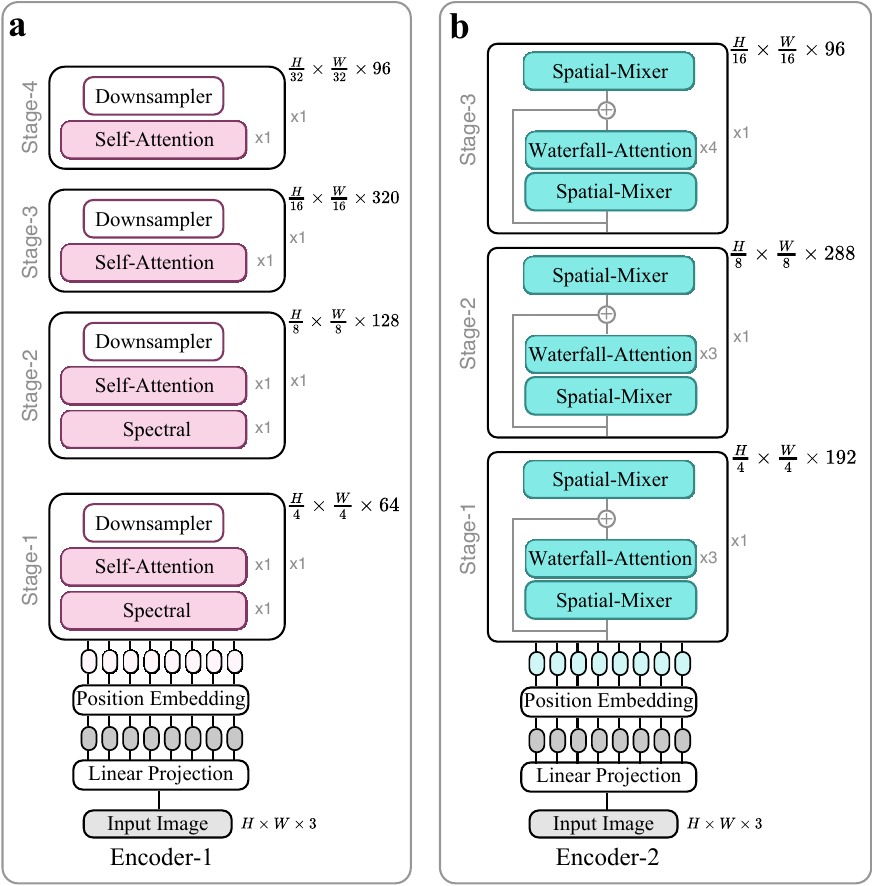} 
\end{center}
\caption{Representation of the architecture of the main components of Tiny-BioMoE: (a) Encoder-1 and (b) Encoder-2.}
\label{encoders}
\end{figure}

Built upon hierarchical vision-transformer principles, incorporates mechanisms that enhance both efficiency and  speed.
\textit{Encoder-2} consists of two core components: the \textit{Spatial-Mixer} and the \textit{Waterfall-Attention} module. The architecture is organized with \textit{Waterfall-Attention} at its center, preceded and followed by \textit{Spatial-Mixer} modules. The input image $I$ is first divided into overlapping $16 \times 16$ patches using a projection layer, with each patch embedded into a $d$-dimensional token.
To capture local information, each patch $T$ is processed with depth-wise convolution:
\begin{equation}
Y_c = K_c * T_c + b_c,
\end{equation}
where $T_c$ and $Y_c$ are the input and output of channel $c$, $K_c$ is the channel-specific kernel, and $b_c$ is the corresponding bias.
Following convolution, batch normalization is applied as $Z_c = \text{BN}(Y_c)$, where BN denotes channel-wise normalization with learnable affine parameters.
A feed-forward network (FFN) then facilitates inter-channel communication,
$\Phi^F(Z_{c}) = W_2 \cdot \text{ReLU}(W_1 \cdot Z_{c} + b_1) + b_2$,
where $\Phi^F(Z_{c})$ is the output of the feed-forward network for the input $Z_{c}$.
$W_1$ and $W_2$ are the weight matrices of the first and second linear layers;
$b_1$ and $b_2$ are the bias terms for the first and second linear layers, respectively,
and $\text{ReLU}$ is the activation function.
Encoder-2 uses a single self-attention layer. For each input embedding,
$X_{i+1}=\Phi^A(X_i)$ is computed,
where $X_i$ is the full input embedding for the $i$-th \textit{block}.
The \textit{Waterfall-Attention} module partitions the embedding into $h$ segments, assigns each to a distinct head, and distributes the workload:
\begin{equation}
\widetilde{X}_{ij} = Attn({X}_{ij} W^Q_{ij}, {X}_{ij} W^K_{ij}, {X}_{ij} W^V_{ij}),
\end{equation}
\begin{equation}
\widetilde{X}_{i+1} = \text{Concat}[\widetilde{X}_{ij}]_{j=1:h}W^P_i,
\end{equation}
where each $j$-th head processes $X_{ij}$, the $j$-th segment of the full embedding $X_i$, structured as $[X_{i1}, X_{i2}, \dots, X_{ih}]$.
Projection layers $W^Q_{ij}$, $W^K_{ij}$, and $W^V_{ij}$ map each segment into distinct subspaces, and $W^P_i$ reunites the concatenated outputs.
The waterfall design enriches each subsequent head by residual addition:
$X^{'}_{ij} = X_{ij} + \widetilde{X}_{i(j-1)}$.
Depth-wise convolution on every $Q$ enables the self-attention to fuse global context with local cues.
The model stacks three stages, each one a token layer deep. It halves the spatial resolution at every stage, progressively reducing the number of tokens. The associated embedding widths are $d=192$, $288$, and $96$, and the stages employ $3$, $3$, and $4$ attention heads, respectively.
\hyperref[encoders]{Figure 2(b}) illustrates the architecture of \textit{Encoder-2}.

\subsubsection{Fusion}
Initially, a \textit{LayerNorm} operation is applied to the input image tensor to ensure consistent normalisation before processing. The normalised input is then passed to both \textit{Encoder-1} and \textit{Encoder-2}, producing embeddings $z_1$ and $z_2$, respectively. Each output is further normalised using a second \textit{LayerNorm}, yielding $\hat{z}_1$ and $\hat{z}_2$.
A lightweight gating network is defined as:
\begin{equation}
g(x) = \text{HardTanh}(\text{ELU}(W x)),
\end{equation}
and produces per-channel modulation coefficients $\alpha_1, \alpha_2 \in [0, 1]$. These are applied element-wise to the encoder outputs:
\begin{equation}
z'_1 = \alpha_1 \odot \hat{z}_1, \quad z'_2 = \alpha_2 \odot \hat{z}_2.
\end{equation}
Each encoder produces a $96$-dimensional embedding, and after re-weighting and concatenation, the resulting feature vector becomes $192$-dimensional:
\begin{equation}
z_{\text{cat}} = [z'_1\,\|\,z'_2] \in \mathbb{R}^{192}.
\end{equation}
A final \textit{LayerNorm} is applied to $z_{\text{cat}}$, yielding the unified output used in downstream tasks.

\subsection{Biosignal Pre-processing \& Visualization} 
\label{biosignal_visualization}
\begin{figure*}
\begin{center}
\includegraphics[scale=0.06]{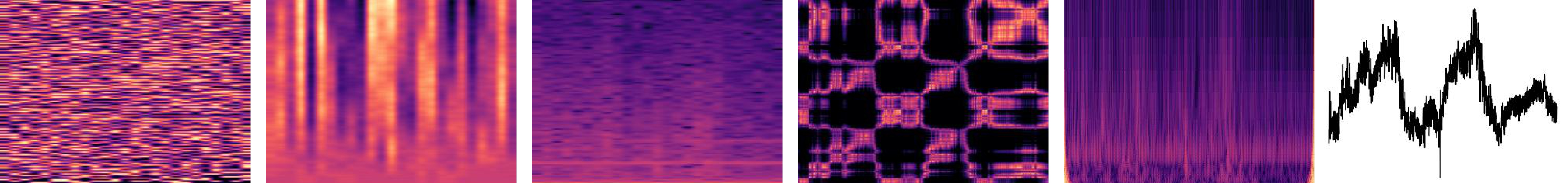} 
\end{center}
\caption{Examples of the six visual representations used for biosignals. From left to right: 
Spectrogram-\textit{Angle}, Spectrogram-\textit{Phase}, Spectrogram-\textit{PSD}, \textit{Recurrence} plot, 
\textit{Scalogram}, and \textit{Waveform}.}
\label{biosignal_samples}
\end{figure*}

The biosignal samples used for both pretraining and pain-related tasks were converted into visual representations to enable compatibility with vision-based models. In total, six distinct transformations were applied:
(1) Spectrogram-\textit{Angle} plots encode the phase angle of the frequency components;  
(2) Spectrogram-\textit{Phase} includes phase information with unwrapping applied to resolve discontinuities;  
(3) Spectrogram-\textit{PSD} displays the power spectral density, reflecting how signal power varies across frequencies and time;  
(4) \textit{Recurrence} plots map the recurrence of states in the signal's phase space to reveal temporal dynamics;  
(5) \textit{Scalograms} represent time-frequency content using continuous wavelet transforms;
(6) \textit{Waveform} diagrams visualize the raw signal over time, capturing its amplitude, frequency, and phase characteristics.
Figure \ref{biosignal_samples} shows one example of each of these six visual representations.
In addition, only the biosignals used in the pain-related tasks \textit{i.e.}, electrodermal activity, blood volume pulse, respiratory signals, and peripheral oxygen saturation were filtered using a low-pass $5$ Hz and band-pass filters of $0.04-1.7$ Hz, $0.05-0.5$ Hz, and  $0.04-1.7$ Hz, respectively.

\subsection{Pretraining}
\label{pretraining}
\textit{Tiny-BioMoE}, the proposed model, serves as an embedding extractor for biosignals. It was trained on $3$ large-scale datasets comprising a total of $4.4$ million samples—details are provided in Table \ref{table:datasets}. These datasets cover a wide range of biosignal modalities, including EEG, EMG, and ECG.
The model was trained using a multi-task learning framework, where each dataset is considered as a distinct supervised task. All $14$ tasks are learned jointly using the following objective:
\begin{equation}
L_{\text{total}} = \sum_{i=1}^{3} \left[ e^{w_i} L_{S_i} + w_i \right],
\end{equation}
where $L_{S_i}$ is the loss associated with the $i$-th task, and $w_i$ are trainable weights that adaptively control each task’s contribution to the overall objective. Training was conducted for $200$ epochs under this setup.

\begin{table}
\caption{Datasets utilized for the multitask learning-based pretraining process of the \textit{PainFormer}.}
\label{table:datasets}
\begin{center}
\begin{threeparttable}
\begin{tabular}{ p{3.0cm} p{2.5cm}  p{1.5cm} }
\toprule
Dataset &Number of Samples  &Modality\\
\midrule
\midrule
\textit{EEG-BST-SZ}      \cite{ford_2013}           &1.20M &EEG\\
\textit{Silent-EMG}      \cite{gaddy_klein_2020}    &1.03M &EMG\\
\textit{ECG HBC Dataset} \cite{kachuee_fazeli_2018} &2.16M &ECG\\\midrule

Total: 3 datasets--tasks &4.39M \\
 
\bottomrule 
\end{tabular}
\begin{tablenotes}
\scriptsize
\item The reported number of samples refers exclusively to the training split, excluding validation and testing sets. Each sample in the corresponding dataset is represented by six distinct visualizations, as described in \ref{biosignal_visualization}.
 
\end{tablenotes}
\end{threeparttable}
\end{center}
\end{table}

\begin{table}
\caption{Number of parameters and FLOPS for the components of the proposed Tiny-BioMoE.}
\label{table:module_parameters}
\begin{center}
\begin{threeparttable}
\begin{tabular}{ P{3.3cm}  P{2.0cm}  P{2.0cm}}
\toprule
Module & Parameters (M) &FLOPS  (G) \\
\midrule
\midrule
Encoder-1   &2.90 &2.36\\
Encoder-2   &4.13 &0.68 \\
\hline
Tiny-BioMoE &7.34 &3.04\\
\bottomrule
\end{tabular}
\begin{tablenotes}
\scriptsize
\item 
\end{tablenotes}
\end{threeparttable}
\end{center}
\end{table}

\subsection{Augmentation Methods \& Regularization}
\begin{figure*}
\begin{center}
\includegraphics[scale=0.20]{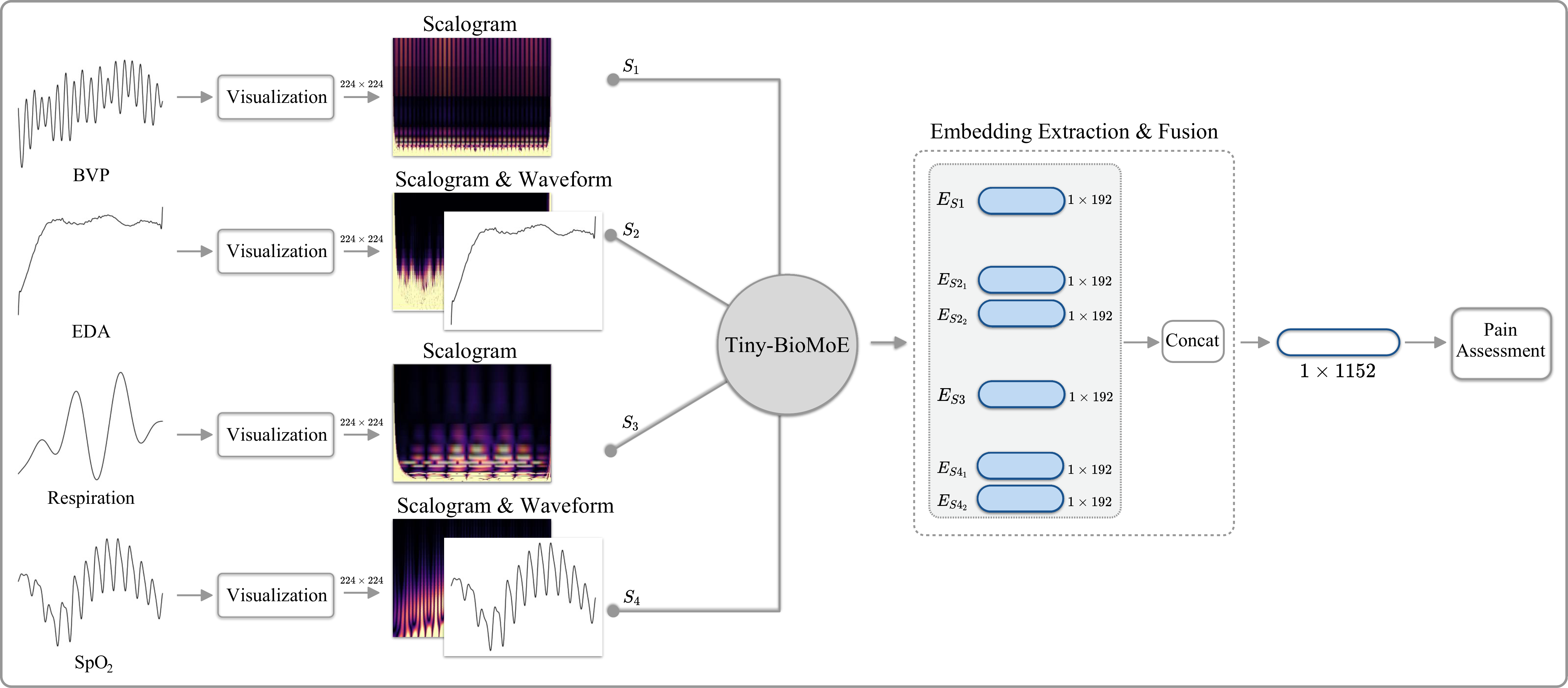} 
\end{center}
\caption{Schematic overview of the proposed pipeline for pain assessment using various modalities and visual representation of them.}
\label{pipeline}
\end{figure*}

Several data augmentation techniques are applied during training for the pain recognition tasks. 
Every $224\times224$ image undergoes a cascade of stochastic transformations.
The \textit{AugMix} method blends three randomly generated augmentation chains with the original image, shifting a mixture of contrast, colour, and geometric perturbations to the image. 
In addition, \textit{TrivialAugment} applies a single randomly chosen operation with a magnitude sampled uniformly. \textit{Centre cropping} is applied with a probability drawn from a given range, where the crop size is selected randomly and the image is resized back to its original dimensions. 
Conditional \textit{Gaussian blurring} applies noise by reducing high-frequency components.
Two \textit{Cutout} masks are applied—one placing a small number of blocks, the other covering the image with a higher number, both using blocks of equal size, $32\times32$.
Beyond data-level augmentation, two stochastic regularizers have been utilized. A \textit{Dropout} layer follows the encoder, with its keep probability gradually reduced linearly over training epochs:
\begin{equation}
p(t) \;=\; p_{\text{start}} + \frac{t}{T}\bigl(p_{\text{end}}-p_{\text{start}}\bigr),\qquad 0\le t\le T.
\end{equation}
The cross-entropy loss employs a \textit{Label-Smoothing} based on the same linear schedule, shifting from a soft target distribution toward one-hot labels.
Finally, a cosine learning-rate profile with \textit{Warmup} and \textit{Cooldown} schedules is also employed.
All stochastic choices---operation selection, magnitudes, cropping ratios, mask locations, and probability draws---are resampled independently for every image at every batch.
Throughout all experiments, the batch size is fixed to $32$ and the learning rate is set to $1\mathrm{e}{-4}$.
Figure \ref{pipeline} presents an overview of the proposed multimodal pipeline.

\section{Experimental Evaluation \& Results}
This study leverages the dataset released by the challenge organizers, which consists of electrodermal activity recordings from $65$ participants. Data collection took place at the Human-Machine Interface Laboratory, University of Canberra, Australia, and is divided into $41$ training, $12$ validation, and $12$ testing subjects. 
Pain stimulation was induced using transcutaneous electrical nerve stimulation (TENS) electrodes positioned on the inner forearm and the back of the right hand. 
Two pain levels were measured: pain threshold---the minimum stimulus intensity perceived as painful (low pain), and pain tolerance---the maximum intensity tolerated before becoming unbearable (high pain). 
The signals have a frequency of $100$ Hz and a duration of approximately $10$ seconds.
We refer to \cite{ai4pain_2025,rojas_hirachan_2023} for a detailed description of the recording protocol and to \cite{ai4pain_2024} for information regarding the previous edition of the challenge.
This study utilises all available modalities---EDA, BVP, respiratory signals, and peripheral oxygen saturation (SpO\textsubscript{2}). All experiments were conducted on the validation subset of the dataset and evaluated within a multi-class classification framework, encompassing three pain levels: No Pain, Low Pain, and High Pain.
The validation results are reported in terms of macro-averaged accuracy, precision, and F1 score. The final results of the testing set are also reported. We note that all experiments followed a deterministic setup, eliminating the effect of random initializations; thus, any performance differences arose strictly from the chosen optimization settings, modalities, or other intentional changes rather than chance.

\subsection{The Impact of Pretraining}
The first series of experiments evaluates the impact of pretraining on model performance. All experiments followed a consistent training setup of $200$ epochs. Table \ref{table:skratch} reports the results obtained with the \textit{Tiny-BioMoE} model trained from scratch, retaining the same architecture but without pretraining. Table \ref{table:pretrained} shows the corresponding results using the pretrained version of \textit{Tiny-BioMoE}, as described in Section \ref{pretraining}, where the model acts as an embedding backbone. In all cases, the model was fine-tuned during training rather than kept frozen.
For the BVP modality, training from scratch resulted in an accuracy of $45.24\%$ with the \textit{Angle} representation, $39.30\%$ for \textit{Phase}, and $42.58\%$ for \textit{PSD}. The \textit{Recurrence} plot reached $42.09\%$, while the highest performance was achieved with the \textit{Scalogram} at $66.53\%$. The \textit{Waveform} also performed reasonably well with $42.90\%$ accuracy.
Using the pretrained model led to improvements across most representations. The most notable improvement was observed for \textit{Recurrence}, which increased by over $5\%$, reaching $67.13\%$.
In the case of EDA, overall performance was higher. With the scratch-trained model, \textit{Angle} achieved $63.02\%$, and \textit{Scalogram} reached $73.41\%$, again emerging as the top representation. Surprisingly, pretraining led to a slight decrease in performance: $61.36\%$ for \textit{Angle} and $71.15\%$ for \textit{Scalogram}. Notably, \textit{PSD} achieved a precision of $80.60\%$, one of the highest observed.
Respiration signals also showed strong results, with $45.61\%$ for \textit{Angle} and $71.91\%$ for \textit{Scalogram}. The pretrained model yielded minor fluctuations, with \textit{Scalogram} improving to $73.53\%$ and \textit{Angle} slightly dropping to $44.78\%$.
For the SpO\textsubscript{2} modality, pretraining resulted in consistent improvements across nearly all representations. \textit{Angle} rose from $51.59\%$ to $55.80\%$, \textit{Phase} from $48.12\%$ to $52.19\%$, and \textit{PSD} from $47.86\%$ to $54.97\%$. The most substantial gains were observed for \textit{Recurrence} and \textit{Waveform}, which increased by $13.46\%$ and $15.62\%$, respectively. The only exception was \textit{Scalogram}, which slightly decreased to $74.31\%$.
Overall, the average accuracy across all representations and modalities increased from $52.13\%$ with the scratch-trained model to $53.96\%$ with the pretrained one. The most significant improvement occurred in the SpO\textsubscript{2} modality, where the average accuracy rose from $55.59\%$ to $62.73\%$.
Note that all the following experiments report results using the pretrained version of \textit{Tiny-BioMoE}.

\begin{table}
\caption{Comparison of performance across different modalities and representations without pretraining (scratch).}

\label{table:skratch}
\begin{center}
\begin{threeparttable}
\begin{tabular}{ P{1.5cm} P{2.0cm} P{1.04cm} P{0.95cm} P{0.95cm}}
\toprule
\multirow{2}[2]{*}{\shortstack{Modality}}
&\multirow{2}[2]{*}{\shortstack{Representation}}
&\multicolumn{3}{c}{Task--MC}\\ 
\cmidrule(lr){3-5} 
 & &Accuracy &Precision &F1\\
\midrule
\midrule
BVP &Angle      &\underline{45.24} &49.42             &\underline{46.72}\\
BVP &Phase      &39.30             &\underline{61.30} &40.48\\
BVP &PSD        &42.58             &58.24             &45.03\\
BVP &Recurrence &42.09             &45.95             &43.47\\
BVP &Scalogram  &\textbf{66.53}    &\textbf{67.33}    &\textbf{66.41}\\
BVP &Waveform   &42.90             &43.71             &41.86\\\midrule

EDA &Angle      &\underline{63.02} &\underline{69.12} &\underline{64.83}\\
EDA &Phase      &53.94             &58.26             &53.21\\
EDA &PSD        &53.06             &43.17             &47.06\\
EDA &Recurrence &59.37             &63.27             &61.11\\
EDA &Scalogram  &\textbf{73.41}    &\textbf{73.59}    &\textbf{73.03}\\
EDA &Waveform   &61.48             &68.32             &63.45\\\midrule

Resp &Angle      &\underline{45.61} &51.85             &45.57\\
Resp &Phase      &40.96             &45.11             &42.85\\
Resp &PSD        &43.42             &\underline{57.49} &\underline{46.08}\\
Resp &Recurrence &36.16             &39.08             &36.22\\
Resp &Scalogram  &\textbf{71.91}    &\textbf{71.89}    &\textbf{71.72}\\
Resp &Waveform   &36.62             &40.20             &32.90\\\midrule

SpO\textsubscript{2} &Angle      &51.59             &58.85             &53.76\\
SpO\textsubscript{2} &Phase      &48.12             &\underline{72.53} &51.26\\
SpO\textsubscript{2} &PSD        &47.86             &67.33             &50.23\\
SpO\textsubscript{2} &Recurrence &\underline{55.18} &59.81             &\underline{57.06}\\
SpO\textsubscript{2} &Scalogram  &\textbf{75.93}    &\textbf{77.24}    &\textbf{75.63}\\
SpO\textsubscript{2} &Waveform   &54.87             &59.74             &56.03\\

\bottomrule 
\end{tabular}
\begin{tablenotes}[para,flushleft] 
\scriptsize                   
\item 
\end{tablenotes}
\end{threeparttable}
\end{center}
\end{table}

\begin{table}
\caption{Comparison of performance across different modalities and representations with the pretrained model.}
\label{table:pretrained}
\begin{center}
\begin{threeparttable}
\begin{tabular}{ P{1.5cm} P{2.0cm} P{1.04cm} P{0.95cm} P{0.95cm}}
\toprule
\multirow{2}[2]{*}{\shortstack{Modality}}
&\multirow{2}[2]{*}{\shortstack{Representation}}
&\multicolumn{3}{c}{Task--MC}\\ 
\cmidrule(lr){3-5} 
 & &Accuracy &Precision &F1\\
\midrule
\midrule
BVP &Angle      &46.06             &49.71             &47.40\\
BVP &Phase      &40.39             &44.43             &41.44\\
BVP &PSD        &43.04             &48.00             &44.42\\
BVP &Recurrence &\underline{47.43} &49.80             &\underline{48.26}\\
BVP &Scalogram  &\textbf{67.13}    &\textbf{65.42}    &\textbf{62.19}\\
BVP &Waveform   &43.67             &\underline{51.28} &44.00\\\midrule

EDA &Angle      &61.36             &63.19             &61.85\\
EDA &Phase      &51.29             &58.22             &53.87\\
EDA &PSD        &52.57             &\textbf{80.60}    &50.82\\
EDA &Recurrence &55.48             &57.05             &56.22\\
EDA &Scalogram  &\textbf{71.15}    &\underline{71.83} &\textbf{69.98}\\
EDA &Waveform   &\underline{63.46} &61.16             &\underline{62.00}\\\midrule

Resp &Angle      &44.78             &45.80             &45.19\\
Resp &Phase      &40.47             &46.28             &41.73\\
Resp &PSD        &\underline{45.29} &\underline{51.40} &\underline{46.18}\\
Resp &Recurrence &38.18             &39.12             &38.47\\
Resp &Scalogram  &\textbf{73.53}    &\textbf{73.93}    &\textbf{73.40}\\
Resp &Waveform   &33.33             &26.78             &29.70\\\midrule

SpO\textsubscript{2} &Angle      &55.80             &58.72             &56.15\\
SpO\textsubscript{2} &Phase      &52.19             &56.37             &53.65\\
SpO\textsubscript{2} &PSD        &54.97             &61.29             &57.14\\
SpO\textsubscript{2} &Recurrence &68.64             &70.45 			   &69.18\\
SpO\textsubscript{2} &Scalogram  &\textbf{74.31}    &\textbf{74.37}    &\textbf{74.25}\\
SpO\textsubscript{2} &Waveform   &\underline{70.49} &\underline{71.66} &\underline{71.03}\\

\bottomrule 
\end{tabular}
\begin{tablenotes}[para,flushleft] 
\scriptsize                   
\item 
\end{tablenotes}
\end{threeparttable}
\end{center}
\end{table}

\subsection{Fusion of Representations}
The next series of experiments focuses on the fusion of representations within each modality. As previously discussed, each biosignal modality includes six distinct visual representations; however, not all contribute equally to performance. The goal is to evaluate which combinations of these representations can enhance the results. 
Fusion is performed by first extracting embeddings for each representation using the pretrained \textit{Tiny-BioMoE}, followed by applying two fusion methods: element-wise addition and concatenation. Two main strategies are examined: (i) fusion of all six representations for each modality, and (ii) fusion of the two top-performing representations per modality. The results are presented in Table \ref{table:fusion_repre}.
For the BVP modality, fusion of all six representations using addition yielded an accuracy of $54.01\%$, while concatenation improved the performance to $58.87\%$, the highest recorded for this modality. In contrast, fusion of only the \textit{Scalogram} and \textit{Recurrence} representations resulted in lower accuracy: $53.08\%$ with addition and $49.18\%$ with concatenation. In all cases, fusion led to decreased performance compared to using only the \textit{Scalogram}, the best standalone representation for BVP.
Regarding the EDA modality, fusion of all representations decreased accuracy to $67.75\%$ with addition and $63.62\%$ with concatenation. However, when only \textit{Scalogram} and \textit{Recurrence} were fused, accuracy improved to $76.85\%$ and $77.88\%$ respectively—both outperforming the best individual representation for this modality.
A similar trend was observed in the respiration signals. Fusion of all representations via concatenation resulted in $69.87\%$, which was lower than the $73.53\%$ achieved using only the \textit{Scalogram}. 
Finally, for the SpO\textsubscript{2} modality, fusion of all six representations did not improve performance. However, fusing \textit{Scalogram} and \textit{Recurrence} via concatenation led to an accuracy of $74.54\%$, showing a slight improvement over the best single-representation result.

\begin{table}
\caption{Comparison of performance across different fusion methods of representations.}
\label{table:fusion_repre}
\begin{center}
\begin{threeparttable}
\begin{tabular}{ P{1.00cm} P{1.8cm} P{0.8cm} P{1.0cm} P{0.90cm} P{0.70cm}}
\toprule
\multirow{2}[2]{*}{\shortstack{Modality}}
&\multirow{2}[2]{*}{\shortstack{Representation}}
&\multirow{2}[2]{*}{\shortstack{Fusion}}
&\multicolumn{3}{c}{Task--MC}\\ 
\cmidrule(lr){4-6} 
& & &Accuracy &Precision &F1\\
\midrule
\midrule

BVP & \makecell{Scal, Recur,\\Angle, Wave\\PSD, Phase} & add
    &\underline{54.01}  &\underline{61.22}  &\underline{56.62}\\\hdashline

BVP & \makecell{Scal, Recur,\\Angle, Wave\\PSD, Phase} & concat
    &\textbf{58.87}  &\textbf{63.40}  &\textbf{60.18} \\\hdashline

BVP & \makecell{Scal, Recur} & add
    &53.08  &54.20  &53.06 \\\hdashline

BVP & \makecell{Scal, Recur} & concat
    &49.18  &61.13  &53.28 \\\midrule

EDA & \makecell{Scal, Wave,\\Angle, Recur\\PSD, Phase} & add
    &67.75  &68.01  &64.63\\\hdashline

EDA & \makecell{Scal, Wave,\\Angle, Recur\\PSD, Phase} & concat
    &63.62  &61.59  &61.64 \\\hdashline

EDA & \makecell{Scal, Recur} & add
    &\underline{76.85}  &\underline{73.70}  &\underline{75.06} \\\hdashline

EDA & \makecell{Scal, Recur} & concat
    &\textbf{77.88}  &\textbf{76.34}  &\textbf{76.88} \\\midrule

Resp & \makecell{Scal, PSD,\\Angle, Phase\\Recur, Wave} & add
    &66.18  &\underline{70.30}  &65.27\\\hdashline

Resp & \makecell{Scal, PSD,\\Angle, Phase\\Recur, Wave} & concat
    &\textbf{69.87}  &67.91  &\underline{68.81} \\\hdashline

Resp & \makecell{Scal, PSD} & add
    &67.36  &\textbf{71.34}  &64.90 \\\hdashline

Resp & \makecell{Scal, PSD} & concat
    &\underline{69.10}  &69.90  &\textbf{69.17} \\\midrule

SpO\textsubscript{2} & \makecell{Scal, Wave,\\Recur, Angle\\PSD, Phase} & add
    &71.96  &\underline{74.59}  &72.84\\\hdashline

SpO\textsubscript{2} & \makecell{Scal, Wave,\\Recur, Angle\\PSD, Phase} & concat
    &\underline{73.84}  &74.11  &\underline{73.79} \\\hdashline

SpO\textsubscript{2} & \makecell{Scal, Recur} & add
    &72.92  &72.93  &72.90 \\\hdashline

SpO\textsubscript{2} & \makecell{Scal, Recur} & concat
    &\textbf{74.54}  &\textbf{74.95}  &\textbf{74.22} \\

\bottomrule 
\end{tabular}
\begin{tablenotes}[para,flushleft] 
\scriptsize                   
\item 
\end{tablenotes}
\end{threeparttable}
\end{center}
\end{table}

\subsection{Fusion of Modalities}
\begin{figure*}
\begin{center}
\includegraphics[scale=0.105]{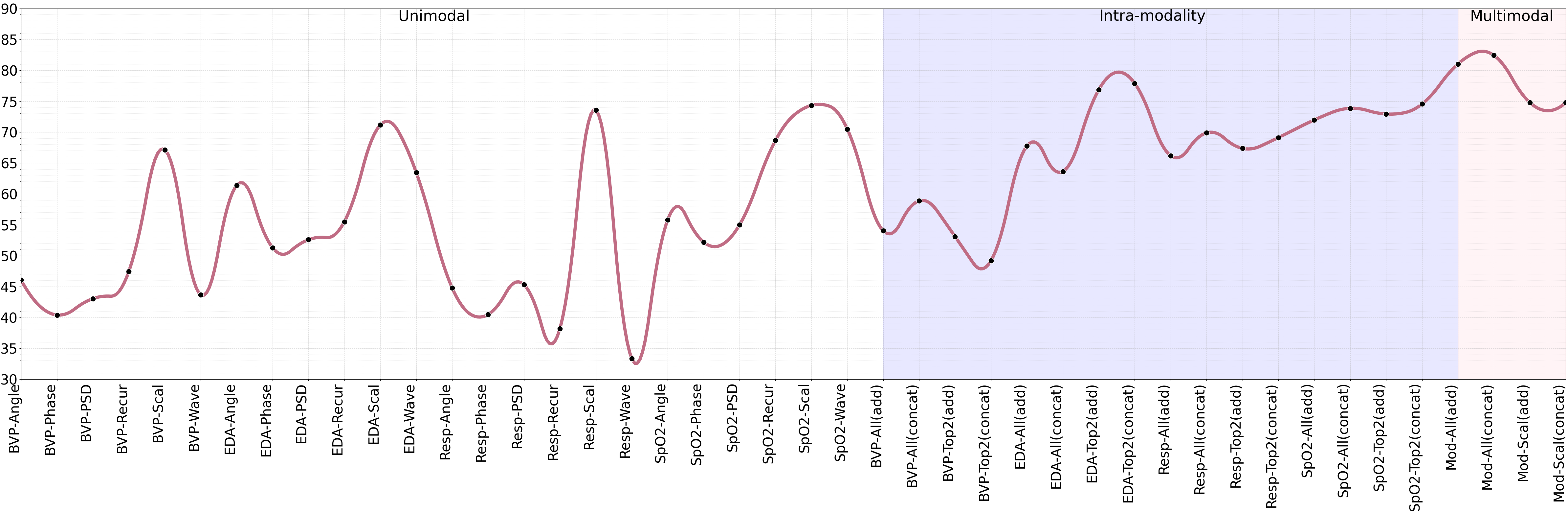} 
\end{center}
\caption{Performance across modality representations, combinations of representations, and combinations of modalities.}
\label{all_results}
\end{figure*}

The final experiments investigate the fusion of modalities along with their corresponding representations. Initially, the best performing representation for each modality is selected and fused using both addition and concatenation methods, as in previous experiments. Additionally, a configuration using only the \textit{Scalogram} representation from each modality is examined, since it generally yields the highest individual performance. The results are shown in Table \ref{table:fusion_modal}.
Specifically, the selected configuration includes the \textit{Scalogram} for BVP, the \textit{Scalogram} and \textit{Waveform} for EDA, the \textit{Scalogram} for respiration, and the \textit{Scalogram} and \textit{Waveform} for SpO\textsubscript{2}. This fusion setup resulted in high performance, achieving $81.02\%$ accuracy with addition and $82.41\%$ with concatenation—the highest scores reported in this study.
Using only the \textit{Scalogram} from each modality for fusion did not further improve performance, but still yielded strong results, with both fusion methods reaching an accuracy of $74.77\%$.
Figure \ref{all_results} visualizes the accuracy performance across individual modality representations, fused representations, and multimodal combinations reported in this study.

\begin{table}
\caption{Comparison of performance across different fusion methods of modalities.}
\label{table:fusion_modal}
\begin{center}
\begin{threeparttable}
\begin{tabular}{ P{1.00cm} P{1.8cm} P{0.8cm} P{1.0cm} P{0.90cm} P{0.70cm}}
\toprule
\multirow{2}[2]{*}{\shortstack{Modality}}
&\multirow{2}[2]{*}{\shortstack{Representation}}
&\multirow{2}[2]{*}{\shortstack{Fusion}}
&\multicolumn{3}{c}{Task--MC}\\ 
\cmidrule(lr){4-6} 
& & &Accuracy &Precision &F1\\
\midrule
\midrule

BVP       & \makecell{Scalogram}        & \multirow{4}{*}{add} & \multirow{4}{*}{\underline{81.02}} & \multirow{4}{*}{\underline{81.72}} & \multirow{4}{*}{\underline{81.36}}\\
EDA       & \makecell{Scal, Wave}  &&&&\\
Resp      & \makecell{Scalogram}        &&&&\\
SpO\textsubscript{2} & \makecell{Scal, Wave} &&&&\\\hdashline

BVP       & \makecell{Scalogram}        & \multirow{4}{*}{concat} & \multirow{4}{*}{\textbf{82.41}} & \multirow{4}{*}{\textbf{84.25}} & \multirow{4}{*}{\textbf{82.43}}\\
EDA       & \makecell{Scal, Wave}  &&&&\\
Resp      & \makecell{Scalogram}        &&&&\\
SpO\textsubscript{2} & \makecell{Scal, Wave} &&&&\\\midrule

BVP       & \makecell{Scalogram}  & \multirow{4}{*}{add} & \multirow{4}{*}{74.77} & \multirow{4}{*}{75.47} & \multirow{4}{*}{74.84}\\
EDA       & \makecell{Scalogram}  &&&&\\
Resp      & \makecell{Scalogram}  &&&&\\
SpO\textsubscript{2} & \makecell{Scalogram} &&&&\\\hdashline

BVP       & \makecell{Scalogram}  & \multirow{4}{*}{concat} & \multirow{4}{*}{74.77} & \multirow{4}{*}{75.75} & \multirow{4}{*}{75.23}\\
EDA       & \makecell{Scalogram}  &&&&\\
Resp      & \makecell{Scalogram}  &&&&\\
SpO\textsubscript{2} & \makecell{Scalogram} &&&&\\

\bottomrule 
\end{tabular}
\begin{tablenotes}[para,flushleft] 
\scriptsize                   
\item 
\end{tablenotes}
\end{threeparttable}
\end{center}
\end{table}

\section{Comparison with Existing Methods}
In this section, the proposed approach is compared with previous studies using the testing set of the \textit{AI4PAIN} dataset. Some of these studies were conducted as part of the \textit{First Multimodal Sensing Grand Challenge}. In contrast, others, including the present work, utilized data from the \textit{Second Multimodal Sensing Grand Challenge}. The key distinction between the two challenges lies in the set of available modalities.
Studies employing facial video or fNIRS have reported strong results, with accuracies of $49.00\%$ by \cite{prajod_schiller_2024} and $55.00\%$ by \cite{nguyen_yang_2024}, respectively. Combining these two modalities also yielded competitive results, although not substantially better than using each modality alone. For instance, \cite{vianto_2025} reported $51.33\%$, and \cite{gkikas_rojas_painformer_2025} achieved $55.69\%$ using fused video and fNIRS data.
Regarding the physiological modalities available in the \textit{Second Grand Challenge}, some approaches achieved notably high performance, while others produced more limited results.
In \cite{gkikas_kyprakis_eda_2025}, an accuracy of $55.17\%$ was reported using EDA in isolation, whereas \cite{gkikas_kyprakis_resp_2025} achieved $42.17\%$ using only the respiration signal.
The proposed method, which combines all available modalities---EDA, BVP, respiration, and SpO\textsubscript{2}—and utilizes specific visual representations for each, achieved an accuracy of $54.89\%$. This ranks among the highest performances reported on the dataset.

\begin{table}
\caption{Comparison of studies on the testing set of the \textit{AI4Pain} dataset.}
\label{table:ai4pain_test}
\begin{center}
\begin{threeparttable}
\begin{tabular}{P{0.7cm} P{1.5cm} P{2.0cm} P{1.5cm} P{1.0cm}}
\toprule
Study & Modality & ML &Parameters & Acc (\%) \\
\midrule
\midrule
\cite{khan_aziz_2025}$^\dagger$                &fNIRS        &ENS             &--    &53.66\\ \hdashline
\cite{nguyen_yang_2024}$^\dagger$              &fNIRS        &Transformer     &--    &55.00\\ \hdashline
\cite{prajod_schiller_2024}$^\dagger$          &Video        &2D CNN          &--    &49.00\\ \hdashline
\cite{gkikas_tsiknakis_painvit_2024}$^\dagger$ &Video, fNIRS &Transformer     &32.92 &46.67\\ \hdashline
\cite{vianto_2025}$^\dagger$                   &Video, fNIRS &CNN-Transformer &86.73 &51.33 \\\hdashline
\cite{gkikas_rojas_painformer_2025}$^\dagger$  &Video, fNIRS &Transformer     &29.45 &55.69 \\\midrule

\cite{gkikas_kyprakis_eda_2025}$^\ddagger$     &EDA         &Transformer &19.60 &55.17\\\hdashline
\cite{gkikas_kyprakis_resp_2025}$^\ddagger$    &Respiration &Transformer &3.62 &42.24 \\\hdashline

Our$^\ddagger$    &EDA, BVP, Resp, SpO$_2$ &MoE    &7.34  &54.89\\

\bottomrule 
\end{tabular}
\begin{tablenotes}
\scriptsize
\item \textbf{ENS}: Ensemble Classifier 
$\pmb{\dagger}$: AI4PAIN-First Multimodal Sensing Grand Challenge $\pmb{\ddagger}$: AI4PAIN-Second Multimodal Sensing Grand Challenge
\end{tablenotes}
\end{threeparttable}
\end{center}
\end{table}

\section{Conclusion}
This study presents our contribution to the \textit{Second Multimodal Sensing Grand Challenge for Next-Generation Pain Assessment (AI4PAIN)}, where all available modalities—EDA, BVP, respiration, and SpO\textsubscript{2} were utilized. We introduced \textit{Tiny-BioMoE}, a lightweight embedding model for biosignal analysis. With only $7.3$ million parameters, a computational cost of $3.04$ GFLOPs, and pretrained on $4.4$ million biosignal-based representations, it serves as an efficient and versatile solution for a wide range of physiology-related tasks.
Evaluated on the \textit{AI4PAIN} dataset for pain recognition, \textit{Tiny-BioMoE} demonstrated strong performance across modalities. A variety of visual representations were also explored. Results showed that the pretrained version of \textit{Tiny-BioMoE} consistently outperformed its non-pretrained counterpart, especially in cases where the individual modality alone had limited discriminative power. This is particularly important, as it enables the use of modalities that might otherwise be disregarded.
A comprehensive set of experiments revealed a clear trend in performance, progressing from isolated representations to intra-modality fusion and ultimately to multimodal combinations. The highest accuracy was achieved through a multimodal approach, combining all four modalities and selected visual representations.
We argue that small, efficient pretrained models, such as \textit{Tiny-BioMoE}, are highly valuable and that future efforts should prioritize their development and open distribution to help democratize access to advanced physiological modeling, regardless of hardware constraints.

\section*{Safe and Responsible Innovation Statement}
This work relied on the \textit{AI4PAIN} dataset \cite{rojas_hirachan_2023,ai4pain_2024,ai4pain_2025}, made available by the challenge organizers, to assess automatic pain recognition methods. All participants confirmed the absence of neurological or psychiatric conditions, unstable health issues, chronic pain, or regular medication use during the session. Before the experiment, participants were thoroughly informed of the procedures, and written consent was obtained. The original study's human-subject protocol received ethical clearance from the University of Canberra's Human Ethics Committee \textit{(approval number: 11837)}. 
The proposed method was developed for continuous pain monitoring, aiming to enhance pain assessment protocols and improve patient care. However, as validation and testing were performed on controlled laboratory data, its deployment in real-world clinical settings requires further investigation and comprehensive evaluation.

\section*{Acknowledgements}
This paper is supported by the projects that have received funding from the
European Union's Horizon 2020 research and innovation programme under grant agreement
$101080905$ (\textit{STRATIFYHF project}).

\bibliographystyle{ACM-Reference-Format}
\bibliography{library}

\end{document}